%% file: mainArXiv.tex
\documentclass{article}
\usepackage{arxiv}
\usepackage{amsfonts,booktabs,multirow,lineno,hyperref,xcolor,amsmath}
\usepackage{blkarray, bigstrut,pifont}
\usepackage{booktabs}
\usepackage{amsthm,subfigure}
\usepackage{hyperref,booktabs}
\usepackage{algorithmic}
\usepackage{algorithm}

\usepackage[utf8]{inputenc} 
\usepackage[T1]{fontenc}    
\usepackage{hyperref}       
\usepackage{url}            
\usepackage{booktabs}       
\usepackage{amsfonts}       
\usepackage{nicefrac}       
\usepackage{microtype}      
\usepackage{lipsum}
\usepackage{graphicx}
\usepackage[shortlabels]{enumitem}
\usepackage{graphicx,amssymb,url,amsfonts,amsmath,mathrsfs,epsf,subfigure}
\usepackage{algorithm,booktabs,epstopdf,hyperref,pifont}
\usepackage{algorithmic}
\usepackage{threeparttable,lineno,multicol,comment}
\usepackage{rotating} 
\usepackage{amssymb}
\usepackage{amsfonts}
\usepackage{siunitx}
\usepackage{url}
\usepackage{eurosym}
\usepackage{courier}
\usepackage{color}
\usepackage{float}
\usepackage{setspace}
\usepackage{diagbox}
\usepackage{multicol}
\usepackage{multirow}
\usepackage{float}
\usepackage{booktabs}

\newcommand{\bm}[1]{\boldsymbol{#1}}

\usepackage{amsmath,amssymb,bm}
\usepackage{caption}
\usepackage{subfigure}
\usepackage{multirow}
\usepackage[english]{babel}
\usepackage{amsthm}

\title{A predict-and-optimize approach to profit-driven churn prevention}

\author{
Nuria G\'omez-Vargas\\
Department of Statistics and Operations Research, Faculty of Mathematics, University of Seville\\
\& Instituto de Matem\'aticas de la Universidad de Sevilla (IMUS), Seville\\
\texttt{ngvargas@us.es}
\And
Sebasti\'an Maldonado\\
Department of Management Control and Information Systems \\
School of Economics and Business\\
University of Chile, Chile\\
\& Instituto Sistemas Complejos de Ingenier\'\i a (ISCI), Chile.\\
\texttt{sebastianm@fen.uchile.cl}
\And
Carla Vairetti \\
Universidad de los Andes, Chile \\
Facultad de Ingenier\'{i}a y Ciencias Aplicadas\\
\& Instituto Sistemas Complejos de Ingenier\'\i a (ISCI), Chile.\\
\texttt{cvairetti@uandes.cl}
}

\begin{document}
\maketitle

\begin{abstract}
In this paper, we introduce a novel predict-and-optimize method for profit-driven churn prevention. We frame the task of targeting customers for a retention campaign as a regret minimization problem. The main objective is to leverage individual customer lifetime values (CLVs) to ensure that only the most valuable customers are targeted. In contrast, many profit-driven strategies focus on churn probabilities while considering average CLVs. This often results in significant information loss due to data aggregation. Our proposed model aligns with the guidelines of Predict-and-Optimize (PnO) frameworks and can be efficiently solved using stochastic gradient descent methods. Results from 12 churn prediction datasets underscore the effectiveness of our approach, which achieves the best average performance compared to other well-established strategies in terms of average profit.
\footnote{This is a preprint of a work under submission and thus subject to change. Changes resulting from the publishing process, such as editing, corrections,structural formatting, and other quality control mechanisms may not be reflected in this version of the document.}
\end{abstract}

\section{Introduction}\label{sec:Intr}
\input{introduction}

\section{Prior work}\label{sec:Prior}
\input{prior_work}

\section{Proposed framework}
\input{proposed_framework}\label{sec:Frame}

\section{Experimental Results}\label{sec:Exp}
\input{experimental_results}

\section{Conclusions}\label{sec:Conc}
\input{conclusions}

\bibliographystyle{IEEEtran}
\bibliography{biblio}

\end{document}

%% file: introduction.tex
The prevailing landscape in the service industry, marked by intense rivalry and well-established markets, necessitates that companies foster robust relationships with their clientele \cite{maldonado2021profit,tungare2023net}. In this sense, retention campaigns that target possibly dissatisfied customers in risk of attrition has become one of the most important marketing actions designed to increase the customer lifetime value (CLV). According to \cite{tungare2023net}, ``loyalty leaders grow revenues roughly 2.5 times as fast as their industry peers and deliver two to five times the shareholder returns over the next ten years''. 

Churn prediction, also referred as defection detection or retention modeling \cite{fujo2022customer,Neslin2006}, has become one of the main marketing analytics applications \cite{verbeke2017profit}. The goal is to identify the customers that are more likely to leave the company via predictive models. This topic has been studied in a wide variety of industries, including the financial sector \cite{maldonado2021profit}, telecommunications \cite{verbraken2012novel}, o higher education (as student dropout) \cite{MALDONADO2021113493}. 

In order to go beyond \emph{churn prediction} and do \emph{churn prevention}, we need to focus on the actions, i.e. the retention campaign, rather than simply identifying potential churners \cite{DEVRIENDT2021497}. One approach is to use profit metrics for model evaluation \cite{maldonado2021profit,verbraken2012novel} or training \cite{LOPEZ2019190,MALDONADO2020273,stripling2018profit}, computing the total or the average profit of the retention campaign instead of utilizing statistical measures.

Traditional profit metrics usually consider an average CLV for all customers to ease the analysis. This simplification, however, can lead to suboptimal decisions in case the CLVs are very heterogeneous, which is common in several industries \cite{maldonado2021profit,Oskarsdottir2018}. To overcome this issue, some metrics compute segment-wise CLVs to account for multiple customer segments \cite{maldonado2021profit}. 

In this study, we go beyond group-wise analysis and propose a predict-and-optimize (PnO) approach to churn prediction. This is an emerging trend in machine learning, garnering significant attention recently \cite{vanderschueren2022predict}. Traditional machine learning models statistical loss functions like mean squared error or cross-entropy, predictions derived from these models might not always lead the optimizer towards the most favorable decisions. In contrast, PnO centers around finding parameters which subsequently feed into an optimization problem. Based on these estimations, the optimization algorithm then determines the best course of action \cite{elmachtoub2022smart,vanderschueren2022predict}. 

In the proposed PnO model, called PnO for churn prevention (PnO$_{\text{cp}}$), the customers are evaluated at an individual level, taking full advantage of the available information and avoiding aggregations. A regret minimization problem is proposed, which can be solved efficiently via stochastic gradient descent. This is the first PnO approach to churn prevention, to the best of our knowledge.

A notable strength of the proposed method is its capacity to explicitly integrate indirect costs into the model. By minimizing the gap between optimal decisions and those driven by the predictive model, the model can be refined to avoid costly errors, such as overlooking potential churners. In comparison, the conventional profit-driven churn prediction framework for model evaluation operates on the presumption that failing to target potential churners incurs no cost, given that it does not result in direct monetary expenditures \cite{maldonado2021profit,verbraken2012novel}.

The remainder of this paper is structured as follows. Section \ref{sec:Prior} presents the relevant background for this study, including literature reviews on profit-based churn prediction and the predict-and-optimize framework. The proposed PnO model for churn prediction is formalized  in Section \ref{sec:Frame}. Experiments on 12 real-world dataset from a Chilean mutual fund company are reported in Section \ref{sec:Exp}, analyzing also the sensitivity of the retention incentive, which is the most relevant parameter. Finally, the main conclusions are presented in Section \ref{sec:Conc}, including also the limitations of the study and  potential future work.

%% file: prior_work.tex
This section is structured as follows: the profit-driven churn prediction framework is introduced in Section \ref{sec:churnpred}, while the predict-and-optimize approach is described in Section \ref{sec:PnO}.

\subsection{Profit-driven churn prediction}\label{sec:churnpred}

The prediction of customer churn has been usually done via binary classification to identify whether a customer will leave the company in the following period \cite{Neslin2006}. Traditional machine learning methods such as ensemble methods or neural networks have been considered for this task \cite{maldonado2021profit,vanderschueren2022predict,verbraken2012novel}. Alternatively, recent approaches include deep learning \cite{fujo2022customer} or social network analysis \cite{ahmad2019customer}. Following the notation proposed in \cite{Verbeke2012,verbraken2012novel}, a customer $i$, described by a vector of variables $\textbf{x}^i$, can be either a churner ($y^i=0$) or a non-churner ($y^i=1$). Notice that it is common in other machine learning studies outside the profit-driven classification literature to represent the churners as $y^i=1$.

Machine learning methods estimate a probabilistic outcome $s \in [0,1]$ and use a threshold or cut-off $t$ to classify customers into churners ($s \le t$) or non-churners ($s > t$). Although the majority of the studies on churn prediction consider statistical measures to find the threshold $t$ to define the best classifier in terms of performance, the evaluation can be also done using goal-oriented metrics. In particular, the optimization of $t$ to maximize the average profit of a retention campaign was proposed in \cite{Verbeke2012}. This approach requires a known cost structure for the problem, which is formalized in Table \ref{tab:cost-benefit}.

\begin{table}[ht]
\centering
\begin{tabular}{@{}cc|cc@{}}
\multicolumn{1}{c}{} &\multicolumn{1}{c}{} &\multicolumn{2}{c}{Predicted} \\ 
\multicolumn{1}{c}{} & 
\multicolumn{1}{c|}{} & 
\multicolumn{1}{c}{$\hat{y}=0$} & 
\multicolumn{1}{c}{$\hat{y}=1$} \\ 
\cline{2-4}
\multirow[c]{2}{*}{\rotatebox[origin=tr]{90}{Actual}}
& $y = 0$ & $f+\gamma\cdot(d - CLV)$ & 0 \\[1.5ex]
& $y=1$ & $f + d$ & 0 \\ 
\cline{2-4}
\end{tabular}
\caption{Classification costs associated to the traditional profit-driven churn prediction framework \cite{verbraken2012novel}.} 
\label{tab:cost-benefit}
\end{table}

The profit-driven framework for churn prediction presented in Table \ref{tab:cost-benefit} considers a retention incentive $d$ which is only incurred in case the offer is accepted. Additionally, the cost of contacting a customer is $f$. In case a would-be churner accepts the offer, the company gains its $CLV$, which is significantly larger than $f+d$. However, it is assumed that only a fraction $\gamma$ of the contacted would-be churners accepts the incentive and stay in the company, while the fraction $1-\gamma$ leaves the company without accepting $d$ despite being identified correctly as churners \cite{verbraken2012novel}. Based on these inputs, the average profit of the retention effort is given by:
\begin{equation}
\label{eq:P1}
{\small 
\begin{split} P_\mathcal{C}(t;\gamma,CLV,\delta,\phi) = CLV \left(\gamma (1-\delta)-\phi\right)\pi_{0} F_{0}(t)- CLV (\delta+\phi) \pi_{1} F_{1}(t), \end{split}
}
\end{equation}
\noindent where $\delta = d/CLV$ and $\phi = f/CLV$, while $\pi_{0}$ and $\pi_1$ and $F_{0}(t)$ and $F_1(t)$ represent the prior probabilities and the cumulative density functions at threshold $t$ for churners and non-churners, respectively \cite{verbraken2012novel}. The maximum profit (MP) criterion \cite{Verbeke2012} is a metric maximizes the profit $P(t;\gamma,CLV,\delta,\phi)$ by optimizing $t$, as follows:

\begin{equation}
\text{MP} = \max_{\forall t} P(t;\gamma,CLV,\delta,\phi),
\end{equation}
The MPC is relatively similar to the profit metric proposed by Neslin et al. \cite{Neslin2006}, which considers the total profit of the campaign instead of a customer-level average profit. Alternatively, a stochastic version of the MP, called expected maximum profit (EMP) criterion, was proposed in \cite{verbraken2012novel}. This strategy assumes that $\gamma$ is a beta distributed random variable. 

Many studies underscore the significance of leveraging individual CLVs rather than relying on a singular value, often the average CLV, in several marketing tasks \cite{Oskarsdottir2018}. Although the MPC/EMPC framework accommodates a distribution for the CLV, the use of a single threshold inevitably leads to an aggregation of the individual values. To overcome this limitation, a multisegment, multithreshold approach was proposed in \cite{maldonado2021profit}. Given $q$ segments of the CLV distribution, obtained via $q$-quantiles, a customer that belongs to the $i$-th CLV segment is classified as a churner if $s \le \tau_i$ ($i=1,...,q$). The average profit for segment-wise retention campaigns is given by:
\begin{equation}
\label{eq:P}
\resizebox{0.9\hsize}{!}{$%
P(\bm{\tau},q;\gamma,\textbf{CLV},\delta,\phi) = \frac{\sum\limits_{i = 1}^q {\left[CLV_i \left(\gamma (1-\delta_i)-\phi_i\right)\pi_{0,i} F_{0,i}(\tau_i)- CLV_i(\delta_i+\phi_i) \pi_{1,i} F_{1,i}(\tau_i)\right]}}{q}
$%
},%
\end{equation}
\noindent where $\pi_{0,i}$ and $\pi_{1,i}$ and $F_{o,i}(t)$ and $F_{1,i}(t)$ represent the segment-level prior probabilities and the cumulative density functions at threshold $t$ for churners and non-churners, respectively \cite{maldonado2021profit}. Additionally, $CLV_i$ is the average CLV for segment $i$, while $\bm{\tau}$ and $\textbf{CLV}$ denote the vectors of thresholds and average per-segment CLVs, respectively. The proposed maximum segment profit (MSP) criterion results from maximizing Eq. \eqref{eq:P} with respect to $\tau_i$, as follows:

\begin{equation}
\text{MSP} = \max_{\forall \bm{\tau},q} P(\bm{\tau};\gamma,\textbf{CLV},\delta,\phi).
\end{equation}

Finally, some approaches have incorporated the MP/EMP framework in the model training, defining cost-sensitive loss functions. For example, ProfLogit \cite{stripling2018profit} implements a regularized logistic regression that maximizes the EMP using a genetic algorithm. Alternatively, a robust optimization approach for profit-driven maximum-margin classification was proposed in \cite{MALDONADO2020273}. However, these approaches have the same limitation as the MP/EMP framework in the sense that they are unable to deal with individual CLVs adequately because a single threshold is considered. The proposed method overcomes this limitation by making individual-level decisions regarding the retention campaign, resulting in the first profit-driven approach that can deal with heterogeneous CLVs during model training.

As highlighted in the introduction, within the MP/EMP framework, the cost associated with mistakenly identifying a potential churner is zero (see Table \ref{tab:cost-benefit} where $y=0$ and $\hat{y}=1$). This is because the computation of the average campaign profit only encompasses actions associated with targeted customers \cite{maldonado2021profit,verbraken2012novel}. The rationale behind this omission of indirect costs is the belief that predictive solutions committing fewer errors with potential churners will be favored when comparing multiple models. Nevertheless, we believe that including these errors in the framework can lead to better decisions, a strength of our suggested PnO approach.

\subsection{Predict-and-optimize}\label{sec:PnO}

Many real world problems frequently confront the task of decision-making under uncertainty, such as planning a retention campaign with no actual knowledge of customers' churn intention. This setting can be described by a parametric optimization problem $\mathcal{P}(y): \min_{z \in \mathcal{Z}} c\bigl(z,y\bigr)$, where $z \in \mathbb{R}^d$ are the decision variables, $y \in \mathbb{R}^d$ are the problem parameters describing the objective function, and $\mathcal{Z} \subset \mathbb{R}^d$ is a nonempty, compact and convex set representing the feasible region (fixed and known with certainty) \cite{elmachtoub2022smart}.

Predict-and-optimize, also known as smart predict-then-optimize, end-to-end learning or decision-focused learning, is a learning paradigm of growing attention within prescriptive analytics, which focuses on integrating the training of the machine learning model that predicts the uncertain parameters and the optimization of the decisions in a single step \cite{elmachtoub2022smart}. Given an output  $\hat{y} = \hat{y}_\theta(\textbf{x})$ of a predictive model, with model parameters $\theta$ and input variables $\textbf{x}^i$, a decision $z^*(\hat{y})$ is implemented by solving $\mathcal{P}(\hat{y})$. Inferring the parameters is an intermediate step of the integrated approach, and the accuracy of $\hat{y}$ is not the primary focus in training. The focus is rather on the error incurred after optimization, when the \textit{prescribed decision} $z^*(\hat{y})$ is implemented, and the cost incurred is with respect to the parameter $y$ that is actually realized. The excess cost due to the fact that $z^*(\hat{y})$ may be suboptimal with respect to $y$ is then $L\bigl(y^i, \hat{y}^i\bigr) = c\bigl(z^*(\hat{y}),y\bigr) - c\bigl(z^*(y) ,y\bigr)$. The latter expression is known as the \textit{suboptimality gap} or \textit{regret}, and its minimization is the criteria by which the model should be trained \cite{mandi2023decisionfocused}.

As we have observed, the loss function is contingent on the solution of an optimization model. According to a recent survey on decision-focused learning \cite{mandi2023decisionfocused}, a key challenge in the integration of prediction and optimization is the differentiation through the optimization problem. When computing $\frac{\partial L}{\partial \theta}$, we need to obtain the term $\frac{\partial z^*(\hat{y})}{\partial \hat{y}}$ within the chain rule. However, $z^*(\hat{y})$ may lack a closed-form representation. An additional challenge arises from decision models operating on discrete variables, which produce discontinuous mappings or sparse (non-informative) gradients. Therefore, examining smooth surrogate models along with their differentiation has been the focus of various works throughout the literature using different methodologies:
analytical differentiation of optimization mappings \cite{agrawal2019differentiable,amos2017optnet}, analytical smoothing of optimization mappings \cite{blondel2020fast,mandi2020interior,wilder2019melding}, smoothing by random perturbations \cite{poganvcic2019differentiation,sahoo2022backpropagation} and differentiation of surrogate loss functions \cite{elmachtoub2022smart,mandi2022decision,mulamba2021contrastive}. Also, there are alternatives to gradient-based decision-focused learning, such as decision trees \cite{elmachtoub2020decision}
.

Yet, to our knowledge, this framework has not been applied specifically to churn prevention. Utilizing it could improve solution quality, as it allows for both the consideration of individual CLV and the missed profit when a potential churner is overlooked in a retention campaign. As will be shown in the next section, we are able to define this intricate closed-form representation of the solution along with a smooth surrogate for the case of churn prevention.

%% file: proposed_framework.tex
The usual way to proceed in defection detection is the estimation of churn probability and then the decision of an optimal threshold according to maximizing profit (MP) from which to target customers for the retention campaign \cite{verbraken2012novel}. Nevertheless, customers with large CLVs are more profitable when retained in contrast to those with smaller CLVs. The latter group is not too attractive for customer retention even when their churn probabilities are high \cite{maldonado2021profit}. In this section, we introduce a novel PnO framework, designed to strike an optimal balance between a customer's profitability and their potential to churn, framed as a regret minimization problem.

When considering individual CLVs, the decision on whether or not to include a customer in the campaign should not depend solely on their probability of churn (estimated from their features), but should also incorporate the value of that particular customer. Thus, we can express the defection detection problem as minimizing the cost
\begin{equation}
\begin{aligned}
\min_{z} \quad & \sum_{i} c\bigl(z^i, y^i\bigr)\\
\textrm{s.t.} \quad & z^i \in \{0,1\} \quad \forall i,
\end{aligned}
\label{eq:Pdd}
\end{equation}
with $ c\bigl(z^i, y^i\bigr) = c_{f,d,CLV^i}\bigl(z^i,y^i\bigr) = z^i \cdot \bigl[ f + y^i\cdot d + (1-y^i) \cdot \gamma \cdot (d-CLV^i) \bigr]$. Recall that $y^i=0$ represents churners and $y^i=1$ denotes non-churners. The parameters $f$ and $d$, correspond to the cost of contacting a customer and the monetary cost of the incentive, respectively. $\gamma$ stands for the fraction of would-be churners who accept the incentive and choose to stay with the company. This parameter is considered deterministic, in line with the MP measure proposed in \cite{Verbeke2012}. $CLV^i$ is the individual customer lifetime value, and $z^i$ is the binary decision variable determining whether or not a customer should be targeted for the retention campaign.

 As there are no budget limits or other constraints, we can derive an expression for the optimal solution in terms of the individual $CLV^i$ in the nominal problem. If $f+\gamma \cdot (d-CLV^i ) \geq 0$, this means the customer is not valuable enough to cover campaign expenses, then $z^{*i} = 0$ (i.e., do not target for the campaign), regardless of whether the client wants to churn or not.  On the other hand, if the company is interested in retaining that customer (which we consider the general case without loss of generality, because if not, simply exclude those instances from the model), and if the intention to churn $y^i$ is known then $z^{*i} = 1 - y^i$, i.e., target the customer if and only if it has the intention of churn. 

 Given the predictive nature of our task, we are operating in a scenario characterized by uncertainty. To address this, we must train a machine learning model $m_\theta$ defined by its parameters $\theta$ using historical data $\bigl\{(\textbf{x}^i, y^i)\bigr\}_i$. The model provides an output $\hat{y}^i = \hat{y}_\theta(\textbf{x}^i)$, which denotes the prediction score for churning of individual $i$. Recall that our prediction does not necessarily need to be a probability (i.e., $\hat{y}^i \in [0,1]$) because it serves merely as an intermediate step. From this, we can derive the explicit relationship between $\hat{y}_\theta(\textbf{x}^i)$ and the prescribed solution $z^*(\hat{y}_\theta(\textbf{x}^i))$:

\begin{equation}
    \begin{aligned}
        z^*(\hat{y}^i) = 1 \quad & \iff \quad f + \hat{y}^i\cdot d + (1-\hat{y}^i) \cdot \gamma \cdot (d-CLV^i) < 0\\
        & \iff \quad f + \hat{y}^i\cdot d + \gamma \cdot (d-CLV^i) - \hat{y}^i \cdot \gamma \cdot (d-CLV^i) < 0\\
        & \iff \quad f + \gamma \cdot (d-CLV^i) < \hat{y}^i \cdot \bigl[ \gamma \cdot (d-CLV^i) - d \bigr]\\
        & \iff \quad \hat{y}^i < \frac{f + \gamma \cdot (d-CLV^i) }{\gamma \cdot (d-CLV^i) - d}
    \end{aligned}
\end{equation}

We denote this middle point as $m^i := \frac{f + \gamma \cdot (d-CLV^i) }{\gamma \cdot (d-CLV^i) - d}$, which is determined by the individual $CLV^i$, and we have the definition of $z^*(\hat{y}_\theta(\textbf{x}^i))$ as a step function:

\begin{equation}
z^*(\hat{y}_\theta(\textbf{x}^i)) =
    \begin{cases}
        1 & \text{if } \hat{y}_\theta(\textbf{x}^i) < m^i\\
        0 & \text{otherwise}
    \end{cases}
    \label{eq:stepZ}
\end{equation}

In this context, for the case of defection prevention, we can provide an explicit definition for the regret:
\begin{equation}
\begin{aligned}
    L(y^i,\hat{y}^i) & = c\bigl(z^*(\hat{y}^i),y^i\bigr) - c\bigl(z^*(y^i) ,y^i\bigr) \\
    & = \bigl(z^*(\hat{y}^i) - 1+y^i\bigr) \cdot \bigl[ f + y^i\cdot d + (1-y^i) \cdot \gamma \cdot (d-CLV^i) \bigr]
\end{aligned}
    \label{eq:regretChurn}
\end{equation}

Eq. \eqref{eq:regretChurn} can serve as a loss function for training the predictive model, incorporating the costs associated with implementing the prescribed solutions and addressing the challenge of computing $\frac{\partial z^*(\hat{y})}{\partial \hat{y}}$ (see Section \ref{sec:PnO}). Moreover, it facilitates the inclusion of individual CLVs during the training phase, distinguishing our model from the MSP criterion proposed in \cite{maldonado2021profit}. This approach is the only existing work that segments different $CLVs$ but relies on already trained classifiers and considers profit solely for model evaluation.

\begin{figure}[ht]
    \centering
    \includegraphics[width=0.75\textwidth]{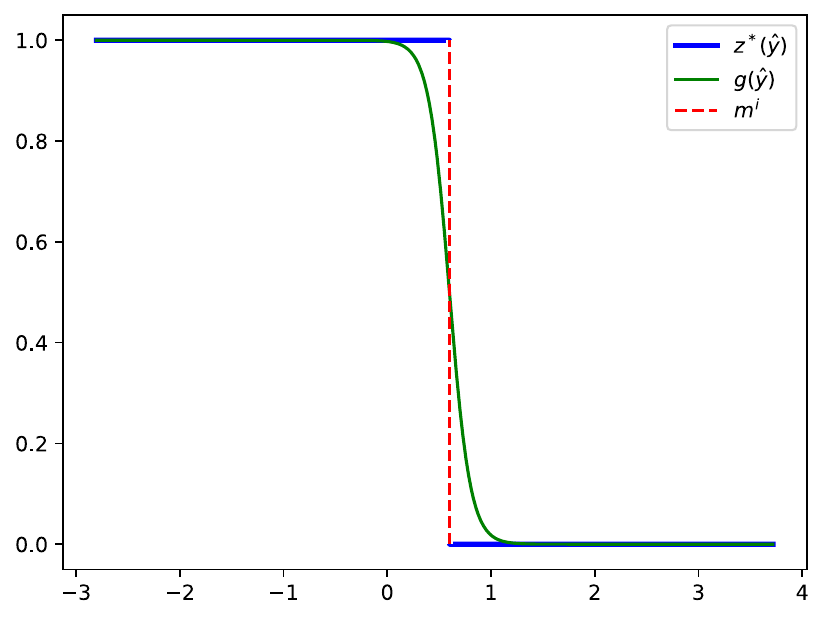}
    \caption{Explicit definition of the prescribed decision $z^*$ as a function of the prediction $\hat{y}$, and the continuous surrogate $g$ as an approximation for $z^*$.}
    \label{fig:z-g}
\end{figure}

Note that the step function in Eq. \eqref{eq:stepZ} is not continuous at $m^i$. As a result, the regret loss function is non-differentiable. However, this discontinuity occurs only at one point, and the function can still be backpropagated using automatic differentiation techniques \cite{Ketkar2021}. On the other hand, the gradient of the loss with respect to the prediction is zero, providing limited guidance to the algorithm. To address this, we define a smooth surrogate for $z^*(\hat{y}_\theta(\textbf{x}^i))$, opting for a continuous relaxation of $z \in [0,1]$, as depicted in Figure \ref{fig:z-g}. This can be achieved using the sigmoid function:
\begin{equation}
g(\hat{y}_\theta(\textbf{x}^i)) = 1 - \sigma\bigl(\hat{y}_\theta(\textbf{x}^i) - m^i\bigr) = 1 - \frac{1}{1 - e^{-s\cdot\bigl(\hat{y}_\theta(\textbf{x}^i) - m^i\bigr)}},
\end{equation}
where we can also vary the slope $s$ --for instance, values $s>1$ for a steeper slope-- as different values might be suited for different problems and different networks \cite{DUBEY202292}, being considered in some cases as a trainable parameter \cite{7410480}.

The continuous surrogate function $g$ can then be used as an alternative version of the regret loss function $L(y,\hat{y})$ defined in Eq. \eqref{eq:regretChurn}, as follows:
\begin{equation}
    L_{\text{smooth}}(y,\hat{y}) = c\bigl(g(\hat{y}),y\bigr) - c\bigl(z^*(y) ,y\bigr).
    \label{eq:Loss_smooth}
\end{equation}

Different machine learning models can serve as functional forms. In this study, we utilize a shallow neural network, deemed appropriate for tabular datasets. Such traditional data structures are most prevalent for churn prediction tasks. However, when more intricate data sources are available, deep learning architectures might be more appropriate. For instance, a recent study indicates that churn prediction data can be presented as a tensor to account for dynamic customer variables, rendering it compatible with deep learning models \cite{mena2023exploiting}. Regarding the optimization process, a stochastic gradient descent procedure is recommended. The Adam optimizer, for example, has emerged as the `de facto' choice for such procedures. For a comprehensive overview of these algorithms, we direct readers to \cite{sun2019survey}.

\subsection{Analyzing the optimality gaps}

As emphasized in Section \ref{sec:churnpred}, the MP/EMP approach does to account for indirect costs related to misclassifying potential churners. When these customers are not targeted, the profit associated with them is zero, as illustrated in Table \ref{tab:cost-benefit}. It might be appropriate to introduce a penalty for missing a would-be churner from whom the company could have garnered their CLV. As we will explore in subsequent lines, this scenario can be aptly addressed using the regret minimization approach.

If the nominal case is $y^i=0$ (churn), it is established that $z^*(y^i) = 1$ (meaning the customer should be targeted for the retention campaign). Consequently, the associated cost is $c\bigl(z^*(y^i), y^i\bigr) = f+\gamma\cdot(d - CLV^i)$. Regarding the prescribed solution, when $z^*(\hat{y}^i)=0$, the cost is $c\bigl(z^*(\hat{y}^i), y^i\bigr) = 0$. This results in a suboptimality gap of
\begin{equation}
L(y^i,\hat{y}^i) = c\bigl(z^*(\hat{y}^i),y^i\bigr) - c\bigl(z^*(y^i) ,y^i\bigr) = -f-\gamma\cdot(d - CLV^i),
\end{equation}
which is strictly greater than zero, hence successfully capturing the indirect cost of losing a potential churner. This approach distinguishes our work from prior profit-driven churn prediction methodologies. Additionally, we effectively incorporate the costs arising from wrongly targeting a non-churner: when $y^i=1$, then $z^*(y^i) = 0$ (i.e., exclude from targeting). However, if the prescribed solution is $z^*(\hat{y}^i) = 1$, then 
\begin{equation}
L(y^i,\hat{y}^i) = f+d - 0 = f +d,
\end{equation}
which is strictly greater than zero and successfully penalizes the incorrect decision. Conversely, when $z^*(\hat{y}^i)=z^*(y^i)$, the optimal solution for the nominal problem  $c\bigl(z^*(\hat{y}^i), y^i\bigr) = c\bigl(z^*(y^i), y^i\bigr)$ is achieved and the regret is zero. These values are summarized in Table \ref{tab:PnO-churn}.

\begin{table}[ht]
    \centering
    \begin{tabular}{@{}cc|cc@{}}
\multicolumn{1}{c}{} &\multicolumn{1}{c}{} &\multicolumn{2}{c}{Prescribed} \\ 
\multicolumn{1}{c}{} & 
\multicolumn{1}{c|}{} & 
\multicolumn{1}{c}{$z^*(\hat{y}^i) = 1$} & 
\multicolumn{1}{c}{$z^*(\hat{y}^i) = 0$} \\ 
\cline{2-4}
\multirow[c]{2}{*}{\rotatebox[origin=tr]{90}{Optimal}}
& $z^*(y^i) = 1$  & 0 & $-f - \gamma \cdot (d-CLV^i)$   \\[1.5ex]
& $z^*(y^i) = 0$  & $f +d$  & 0 \\ 
\cline{2-4}
\end{tabular}
    \caption{Possible values of regret $L(y^i,\hat{y}^i) = c\bigl(z^*(\hat{y}^i),y^i\bigr) - c\bigl(z^*(y^i) ,y^i\bigr)$.}
    \label{tab:PnO-churn}
\end{table}

This approach might appear analogous to the works on instance-dependent cost-sensitive learning \cite{bahnsen2014,hoppner2022instance}, as both share the same goal: to consider the actual business objective in the training process. However, by framing the problem as the solution to an optimization problem, as we do in \eqref{eq:Pdd}, we go beyond incorporating costs into the objective function. This approach also facilitates the imposition of global hard constraints encompassing all instances, thereby refining the feasible region. For instance, this can involve constraints on how many customers can be approached in a campaign. In contrast, cost-sensitive learning might only allow for the inclusion of soft constraints as penalizations in the loss function. 
Furthermore, a study solely based on the cost matrix in Table \ref{tab:cost-benefit}, associated with the traditional profit-driven churn prediction framework, without the comparative analysis as done in the regret loss tailored to churn prevention, would not have allowed us to account for the cost of not making the optimal decision and missing a potential churner, which is also a key aspect of our work. 


%% file: experimental_results.tex
We apply the proposed framework for addressing churn prediction via a PnO model to datasets from a Chilean company. This section is structured as follows: Section \ref{sec:data} offers a detailed description of the dataset; Section \ref{sec:expsetup} delves into the experimental setting; the primary results are encapsulated in Section \ref{sec:summclas}; and lastly, Section \ref{sec:sensanalysis} undertakes a sensitivity analysis of the main parameters within our proposed framework.

\subsection{Data description}\label{sec:data}

The proposed framework is applied to data from a Chilean company which contain 24 features regarding: customer demographics, customer behavior with the company in terms of redemption and repurchase activities, and financial indexes related to the funds themselves (for an extensive description of the datasets features we refer to \cite{maldonado2021profit}). We evaluate 12 different datasets (January -- December, 2017), each being a customer base in which churn is predicted over a three-month period. Individual CLVs are obtained in Chilean Pesos (CLP, CLP/$\textup{\euro}$ exchange rate equals 732.743 on average in 2017). The relevant information for each dataset is summarized in Tables \ref{tab:datasets1}-\ref{tab:datasets2}. 

\begin{table}[ht]
    \centering
    \caption{Numbers of customers (instances), variables, and churn rate for January to June datasets.}
    \label{tab:datasets1}
    \begin{tabular}{ccccccc}
    \toprule
         \multicolumn{1}{l}{} & Jan. & Feb. & Mar. & Apr. & May & Jun.  \\
    \midrule
         Train size & 786 & 792 & 792 & 818 & 844 & 889 \\
         Test size & 197 & 198 & 198 & 205 & 211 & 223  \\
         Churn rate & $16.99\%$ & $16.97\%$ & $17.17\%$ & $16.32\%$ & $17.25\%$ & $18.35\%$ \\
         $\overline{CLV}$ & $85.00\textup{\euro}$ & $85.00\textup{\euro}$ & $88.20\textup{\euro}$ & $88.40\textup{\euro}$ & $86.80\textup{\euro}$ & $91.20\textup{\euro}$ \\
\bottomrule
    \end{tabular}
\end{table}

\begin{table}[ht!]
    \centering
    \caption{Numbers of customers (instances), variables, and churn rate for July to December datasets.}
    \label{tab:datasets2}
    \begin{tabular}{ccccccc}
\toprule
         \multicolumn{1}{l}{} & Jul. & Aug. & Sep. & Oct. & Nov. & Dec. \\
\midrule
         Train size & 924 & 930 & 938 & 961 & 972 & 962\\
         Test size & 231 & 233 & 235 & 241 & 244 & 241 \\
         Churn rate & $19.74\%$ & $18.23\%$ & $19.35\%$ & $20.38\%$ & $21.46\%$ & $17.87\%$ \\
         $\overline{CLV}$ & $91.00\textup{\euro}$ & $92.20\textup{\euro}$ & $90.60\textup{\euro}$ & $87.40\textup{\euro}$ & $87.20\textup{\euro}$ & $89.40\textup{\euro}$ \\
\bottomrule      
    \end{tabular}
\end{table}

For the profit measures, we adhere to the parameters suggested in \cite{Verbeke2012,verbraken2012novel}: the probability parameter $\gamma$ is set to $0.3$ throughout the literature (corresponding to the mean of a beta distribution with shape parameters $\alpha=6$ and $\beta=14$); and the parameter $f$ is set to 1000 CLP ($1.36\textup{\euro}$), which is very close to the value $f=1\textup{\euro}$ in \cite{verbraken2012novel}. Finally, the parameter $d$ is rather complex to set without having information on prior retention campaigns in the mutual fund industry, so we explore a wide range of values $d \in \{\frac{\overline{CLV}}{20}, \frac{\overline{CLV}}{15}, \frac{\overline{CLV}}{10}, \frac{\overline{CLV}}{5}, \frac{\overline{CLV}}{3}\}$ and we present the results for $d=\frac{\overline{CLV}}{20}$, which is the same proportion originally used in \cite{verbraken2012novel}.

\subsection{Experimental setup}\label{sec:expsetup}

We performed a total of 60 experiments combining the 12 datasets with the 5 different values for parameters $d$. We conducted an empirical comparison of the performance of our model following a predict-and-optimize framework for a profit-driven churn prediction (PnO$_{\text{cp}}$) against various well-known classifiers as well as other relevant profit-driven methods. We report the accuracy and the profit of the campaign (considering individual CLVs) computed on the test set (approximately 200 customers for all datasets).

For our methodology, a shallow neural network trained via gradient descent methods is a natural choice. We consider a single hidden layer of size 12, which is half the size of the input variables, and an output layer consisting of one neuron that outputs the churn score. We employed $L_{\text{smooth}}(\hat{y},y)$ (see Eq. \eqref{eq:Loss_smooth}) as the loss function for model training. Monte-Carlo cross-validation was used on the training set to tune the initial learning rate $\lambda \in \{0.01, 0.001, 0.0001\}$ for the Adam algorithm and number of epochs $e \in \{10, 50, 100\}$. For the remaining hyperparameters, we adopt the default settings as described in the PyTorch implementation\footnote{\url{https://pytorch.org/docs/stable/generated/torch.optim.Adam.html}}. For model validation, we used the mean of the loss from Eq. \eqref{eq:Loss_smooth} across 10 different seeds.

In our comparative analysis of profit-driven methods, we evaluated the ProfLogit model \cite{stripling2018profit} using its default configuration, as implemented by the authors, with the number of iterations set to 50. We also assessed the multi-threshold, multi-segment framework from \cite{maldonado2021profit} (MSP) with $q=2$ segments. For the latter metric, we consider the following classifiers: logistic regression (log), random forest (rf), CART decision tree (CART), k-nearest neighbors (KNN) and support vector machine (SVM), consistent with their original configurations reported in \cite{maldonado2021profit}. Our evaluation covers both the performance achieved with these standard classifiers using statistical measures and with the MSP measure. It is important to note that while our model and ProfLogit leverage profit metrics to address the class imbalance issue inherent in the datasets, we still applied the SMOTE oversampling technique \cite{chawla2002smote} for training the other classifiers. This is in line with the experimental setting used in \cite{maldonado2021profit}.

\subsection{Performance summary}\label{sec:summclas}

The performance on the test set for each classification method across the 12 datasets is detailed in Tables \ref{tab:results1}-\ref{tab:results2}. Recall that our optimization problem, as expressed in Eq. \eqref{eq:Pdd}, aims to minimize cost, and then a negative objective function value implies we achieved benefits. Therefore, $-\sum_{i \in \mathcal{D}_\text{test}} c(z^*(y^i),y^i) \geq 0$ is the optimal total profit in the nominal problem over the test set $\mathcal{D}_\text{test}$. These values are reported in the first row of Tables \ref{tab:results1}-\ref{tab:results2}. On the other hand, $-\sum_{i \in \mathcal{D}_\text{test}} c(z^*(\hat{y}^i),y^i)$ denotes the actual profit obtained with a given classifier. By comparing these two measures, we can assess how well each model performs in relation to the optimal decisions. Alongside the profit metric, we also present accuracy for a comprehensive perspective.

\begin{table}[ht!]
    \centering
        \caption{Test performance for all classification methods when incentive $d=\frac{\overline{CLV}}{20}$. Below each month (January to June) we find the optimal profit $-\sum_{i \in \mathcal{D}_\text{test}} c(z^*(y^i),y^i)$ in the test set with the nominal data. For each model, we report on total profit $-\sum_{i \in \mathcal{D}_\text{test}} c(z^*(\hat{y}^i),y^i)$ over test set (accuracy is also reported between brackets).}
    \label{tab:results1}
        \scalebox{0.9}{
    \begin{tabular}{lcccccc}
\toprule
         \multirow{2}{*}{} & Jan. & Feb. & Mar. & Apr. & May & Jun. \\
         & $663.94\textup{\euro}$ & $1015.78\textup{\euro}$ & $1035.85\textup{\euro}$ & $2435.10\textup{\euro}$ & $1133.70\textup{\euro}$ & $2130.97\textup{\euro}$ \\
\midrule
         \multirow{2}{*}{PnO$_{\text{cp}}$} & $\bm{110.12\textup{\euro}}$ & $\bm{502.55\textup{\euro}}$ & $\bm{546.70\textup{\euro}}$ & $\bm{1869.45\textup{\euro}}$ & $25.30\textup{\euro}$ & $1437.83\textup{\euro}$ \\
                              & ($59.9\%$) & ($52.02\%$) & ($58.59\%$) & ($50.24\%$) & ($69.19\%$) & ($44.84\%$) \\
         \multirow{2}{*}{ProfLogit} & $69.31\textup{\euro}$ & $99.81\textup{\euro}$ & $91.17\textup{\euro}$ & $1286.14\textup{\euro}$ & $282.65\textup{\euro}$ & $\bm{1582.02\textup{\euro}}$ \\
                                    & ($\bm{76.14\%}$) & ($72.22\%$) & ($\bm{76.77\%}$) & ($77.07\%$) & ($44.08\%$) & ($\bm{80.27\%}$) \\
        \multirow{2}{*}{MSP$_{\text{KNN}}$} & $-137.28\textup{\euro}$ & $60.02\textup{\euro}$ & $32.96\textup{\euro}$ & $1625.62\textup{\euro}$ & $310.88\textup{\euro}$ & $1513.68\textup{\euro}$ \\
                                    & ($53.3\%$) & ($41.92\%$) & ($46.46\%$) & ($47.32\%$) & ($48.34\%$) & ($55.16\%$) \\
        \multirow{2}{*}{MSP$_{\text{log}}$} & $-49.46\textup{\euro}$ & $25.39\textup{\euro}$ & $366.55\textup{\euro}$ & $1561.37\textup{\euro}$ & $284.68\textup{\euro}$ & $1420.90\textup{\euro}$\\
                                    & ($60.91\%$) & ($41.92\%$) & ($41.41\%$) & ($35.61\%$) & ($39.81\%$) & ($48.43\%$) \\
        \multirow{2}{*}{MSP$_{\text{rf}}$} & $-142.89\textup{\euro}$ & $342.52\textup{\euro}$ & $408.34\textup{\euro}$ & $1753.88\textup{\euro}$ & $\bm{495.31\textup{\euro}}$ & $1438.44\textup{\euro}$\\
                                    & ($52.79\%$) & ($39.39\%$) & ($45.96\%$) & ($45.85\%$) & ($47.87\%$) & ($51.12\%$) \\
        \multirow{2}{*}{MSP$_{\text{CART}}$} & $-62.09\textup{\euro}$ & $168.87\textup{\euro}$ & $349.24\textup{\euro}$ & $1475.59\textup{\euro}$ & $135.15\textup{\euro}$ & $1294.69\textup{\euro}$ \\
                                    & ($58.88\%$) & ($25.25\%$) & ($39.90\%$) & ($19.02\%$) & ($60.19\%$) & ($38.57\%$) \\
        \multirow{2}{*}{MSP$_{\text{SVM}}$} & $47.35\textup{\euro}$ & $-14.42\textup{\euro}$ & $60.38\textup{\euro}$ & $1743.85\textup{\euro}$ & $204.08\textup{\euro}$ & $1346.11\textup{\euro}$ \\
                                    & ($69.54\%$) & ($35.35\%$) & ($41.92\%$) & ($51.22\%$) & ($33.18\%$) & ($42.60\%$) \\
        \multirow{2}{*}{KNN} & $-230.66\textup{\euro}$ & $47.14\textup{\euro}$ & $151.26\textup{\euro}$ & $584.95\textup{\euro}$ & $126.68\textup{\euro}$ & $1517.49\textup{\euro}$ \\
                                    & ($44.16\%$) & ($52.53\%$) & ($58.08\%$) & ($53.17\%$) & ($45.50\%$) & ($59.19\%$) \\
        \multirow{2}{*}{Logistic} & $-54.88\textup{\euro}$ & $76.55\textup{\euro}$ & $-23.96\textup{\euro}$ & $1159.23\textup{\euro}$ & $-103.99\textup{\euro}$ & $1339.84\textup{\euro}$ \\
                                    & ($61.42\%$) & ($71.21\%$) & ($65.66\%$) & ($62.93\%$) & ($61.61\%$) & ($64.57\%$) \\
       \multirow{2}{*}{RF} & $-49.17\textup{\euro}$ & $168.17\textup{\euro}$ & $32.04\textup{\euro}$ & $143.74\textup{\euro}$ & $39.60\textup{\euro}$ & $52.16\textup{\euro}$ \\
                                    & ($74.62\%$) & ($\bm{78.28\%}$) & ($74.75\%$) & ($\bm{80.00\%}$) & ($\bm{77.25\%}$) & ($73.99\%$) \\
        \multirow{2}{*}{CART} & $-139.93\textup{\euro}$ & $64.66\textup{\euro}$ & $43.60\textup{\euro}$ & $51.00\textup{\euro}$ & $26.80\textup{\euro}$ & $221.17\textup{\euro}$ \\
                                    & ($65.99\%$) & ($67.68\%$) & ($68.18\%$) & ($75.12\%$) & ($66.35\%$) & ($67.26\%$) \\
        \multirow{2}{*}{SVM} & $-120.85\textup{\euro}$ & $-2.93\textup{\euro}$ & $114.81\textup{\euro}$ & $1341.47\textup{\euro}$ & $-64.68\textup{\euro}$ & $1306.54\textup{\euro}$ \\
                                    & ($53.3\%$) & ($61.62\%$) & ($65.66\%$) & ($60.49\%$) & ($55.92\%$) & ($60.54\%$) \\
\bottomrule 
    \end{tabular}}
\end{table}

\begin{table}[ht!]
    \centering
    \caption{Test performance for all classification methods when incentive $d=\frac{\overline{CLV}}{20}$. Below each month (July to December) we find the optimal profit $-\sum_{i \in \mathcal{D}_\text{test}} c(z^*(y^i),y^i)$ in the test set with the nominal data. For each model, we report on total profit $-\sum_{i \in \mathcal{D}_\text{test}} c(z^*(\hat{y}^i),y^i)$ over test set (accuracy is also reported between brackets).}
    \label{tab:results2}
    \scalebox{0.9}{
    \begin{tabular}{lcccccc}
\toprule
         \multirow{2}{*}{} & Jul. & Aug. & Sep. & Oct. & Nov. & Dec. \\
         & $1826.48\textup{\euro}$ & $1025.02\textup{\euro}$ & $743.51\textup{\euro}$ & $1462.02\textup{\euro}$ & $1150.08\textup{\euro}$ & $708.94\textup{\euro}$\\
\midrule
         \multirow{2}{*}{PnO$_{\text{cp}}$} & $\bm{1115.31\textup{\euro}}$ & $219.72\textup{\euro}$ & $-12.32\textup{\euro}$ & $\bm{716.48\textup{\euro}}$ & $487.86\textup{\euro}$ & $\bm{85.27\textup{\euro}}$\\
                              & ($45.02\%$) & ($37.34\%$) & ($40.43\%$) & ($65.98\%$) & ($60.25\%$) & ($64.73\%$)\\
         \multirow{2}{*}{ProfLogit} & $266.62\textup{\euro}$ & $272.83\textup{\euro}$ & $9.60\textup{\euro}$ & $424.13\textup{\euro}$ & $\bm{497.84\textup{\euro}}$ & $47.80\textup{\euro}$\\
                                    & ($\bm{74.03\%}$) & ($77.68\%$) & ($74.04\%$) & ($61.41\%$) & ($64.75\%$) & ($72.20\%$)\\
        \multirow{2}{*}{MSP$_{\text{KNN}}$} & $939.89\textup{\euro}$ & $131.30\textup{\euro}$ & $-80.43\textup{\euro}$ & $465.03\textup{\euro}$ & $168.31\textup{\euro}$ & $-111.85\textup{\euro}$\\
                                    & ($38.10\%$) & ($36.91\%$) & ($44.26\%$) & ($27.80\%$) & ($38.93\%$) & ($42.74\%$)\\
        \multirow{2}{*}{MSP$_{\text{log}}$} & $512.24\textup{\euro}$ & $195.37\textup{\euro}$ & $7.87\textup{\euro}$ & $562.44\textup{\euro}$ & $333.32\textup{\euro}$ & $-178.22\textup{\euro}$\\
                                    & ($45.45\%$) & ($66.09\%$) & ($48.51\%$) & ($34.85\%$) & ($51.64\%$) & ($37.34\%$)\\
        \multirow{2}{*}{MSP$_{\text{rf}}$} & $997.90\textup{\euro}$ & $121.78\textup{\euro}$ & $-96.34\textup{\euro}$ & $550.55\textup{\euro}$ & $274.73\textup{\euro}$ & $-282.19\textup{\euro}$\\
                                    & ($41.13\%$) & ($54.94\%$) & ($47.66\%$) & ($39.83\%$) & ($43.03\%$) & ($29.46\%$)\\
        \multirow{2}{*}{MSP$_{\text{CART}}$} & $744.99\textup{\euro}$ & $84.22\textup{\euro}$ & $-191.43\textup{\euro}$ & $361.89\textup{\euro}$ & $117.46\textup{\euro}$ & $-368.00\textup{\euro}$\\
                                    & ($20.78\%$) & ($51.07\%$) & ($32.34\%$) & ($20.33\%$) & ($39.34\%$) & ($32.37\%$)\\
        \multirow{2}{*}{MSP$_{\text{SVM}}$} & $511.23\textup{\euro}$ & $\bm{320.26\textup{\euro}}$ & $86.30\textup{\euro}$ & $539.52\textup{\euro}$ & $388.28\textup{\euro}$ & $-95.63\textup{\euro}$\\
                                    & ($47.19\%$) & ($48.93\%$) & ($56.17\%$) & ($33.20\%$) & ($53.28\%$) & ($42.74\%$)\\
        \multirow{2}{*}{KNN} & $230.69\textup{\euro}$ & $39.89\textup{\euro}$ & $-171.51\textup{\euro}$ & $503.04\textup{\euro}$ & $243.33\textup{\euro}$ & $-107.53\textup{\euro}$\\
                                    & ($57.58\%$) & ($50.21\%$) & ($54.89\%$) & ($50.21\%$) & ($54.10\%$) & ($47.30\%$)\\
        \multirow{2}{*}{Logistic} & $238.37\textup{\euro}$ & $204.53\textup{\euro}$ & $67.08\textup{\euro}$ & $361.82\textup{\euro}$ & $329.35\textup{\euro}$ & $-53.85\textup{\euro}$\\
                                    & ($64.50\%$) & ($63.52\%$) & ($65.11\%$) & ($61.00\%$) & ($64.34\%$) & ($58.92\%$)\\
       \multirow{2}{*}{RF} & $-15.04\textup{\euro}$ & $203.28\textup{\euro}$ & $77.33\textup{\euro}$ & $417.47\textup{\euro}$ & $5.04\textup{\euro}$ & $34.40\textup{\euro}$\\
                                    & ($75.32\%$) & ($\bm{78.97\%}$) & ($\bm{80.43\%}$) & ($\bm{75.93\%}$) & ($\bm{72.54\%}$) & ($\bm{76.35\%}$)\\
        \multirow{2}{*}{CART} & $284.58\textup{\euro}$ & $117.85\textup{\euro}$ & $-50.03\textup{\euro}$ & $377.32\textup{\euro}$ & $260.94\textup{\euro}$ & $-112.77\textup{\euro}$\\
                                    & ($51.52\%$) & ($72.53\%$) & ($65.96\%$) & ($61.41\%$) & ($54.51\%$) & ($69.29\%$)\\
        \multirow{2}{*}{SVM} & $161.96\textup{\euro}$ & $162.21\textup{\euro}$ & $\bm{115.32\textup{\euro}}$ & $562.39\textup{\euro}$ & $359.19\textup{\euro}$ & $-16.17\textup{\euro}$\\
                                    & ($64.94\%$) & ($54.94\%$) & ($65.53\%$) & ($54.36\%$) & ($54.10\%$) & ($57.68\%$)\\
\bottomrule   
    \end{tabular}}
\end{table}

Our experiments reported in Tables \ref{tab:results1}-\ref{tab:results2} show that PnO$_{\text{cp}}$ performs better than the alternative approaches in terms of profit, achieving the most profitable solution for seven of the 12 datasets. This result confirms the virtues of the proposed approach to model profit-based classification problems adequately. For the statistical measure (accuracy), random forest achieves the best overall performance. 

To summarize the results presented in Tables \ref{tab:results1}-\ref{tab:results2}, we rank each classification method based on its relative position across the 12 datasets using the profit in the test set as the main performance measure. A ranking of 1 for a given model and dataset means that it is the top-performing one in terms of this metric. Then, all the rankings are averaged for each technique, obtaining a measure for overall classification performance. This average rank is reported in Table \ref{tabTest} (second column), together with the average profit (third column). 

Additionally, we implement the Friedman test with Iman-Davenport correction and the Holm test, which are well-known approaches used to assess statistical significance in machine learning \cite{Demsar2006}. First, the Friedman with Iman-Davenport correction test is utilized to identify whether there are significant differences between the 12 approaches. This is a non-parametric statistical test used for comparing multiple models across multiple datasets, where the average rankings of the algorithms are considered as a measure of overall performance. The value of this test is 4.1018, rejecting the null hypothesis of similar average ranks with a p-value below 0.0001 and implying that at least one algorithm's performance differs across the datasets. 

The Holm post-hoc test is next implemented to pinpoint which algorithms are different from each other, assessing statistical significance between the top-ranked method (PnO$_{\text{cp}}$, see Table \ref{tabTest}) and the rest of the 11 classifiers. The Nemenyi's Z test performs pairwise comparisons, obtaining a p-value (fourth column in Table \ref{tabTest}), which is compared to a significant threshold $\alpha/(j-1)$, with $j=2,...,12$ and $\alpha=5\%$ (fifth column). The outcome of the test presented in the sixth column is `reject' when its p-value falls below the threshold.

\begin{table}[ht!]
\centering
\caption{Average performance and Holm test for the 12 classifiers on all 12 datasets.}
\label{tabTest}
\begin{tabular}{lccccc}
\toprule
Method & Avg. Rank & Avg. profit($\textup{\euro}$)  & p-value & $\frac{\alpha}{(j-1)}$ & Outcome \\
\midrule
 PnO$_{\text{cp}}$ & 2.7917 & 592.00$\textup{\euro}$ & - & - & - \\
 ProfLogit & 4.4167 & 410.92$\textup{\euro}$ & 0.2696 & 0.0500 &  not reject \\
 MSP$_{\text{SVM}}$ & 5.0833 & 428.00$\textup{\euro}$ & 0.1195 & 0.0250 &  not reject \\
 MSP$_{\text{RF}}$ & 5.4583 & 488.58$\textup{\euro}$ & 0.0700 & 0.0167 &  not reject \\
 MSP$_{\text{log}}$ & 5.5000 & 420.17$\textup{\euro}$ & 0.0658 & 0.0125 &  not reject \\
 MSP$_{\text{KNN}}$ & 6.7500 & 409.92$\textup{\euro}$ & 0.0072 & 0.0100 & reject \\
 SVM & 6.7917 & 326.50$\textup{\euro}$ & 0.0066 & 0.0083 & reject \\
 Logistic & 7.7083 & 295.00$\textup{\euro}$ & 0.0008 & 0.0071 & reject \\
 RF & 7.8750 & 92.33$\textup{\euro}$ & 0.0006 & 0.0063 & reject \\
 MSP$_{\text{CART}}$ & 8.1250 & 342.58$\textup{\euro}$ & 0.0003 & 0.0056 & reject \\
 KNN & 8.3333 & 244.42$\textup{\euro}$ & 0.0002 & 0.0050 & reject \\
 CART & 9.1667 & 95.50$\textup{\euro}$ & 0.0000 & 0.0045 & reject \\
\bottomrule
\end{tabular}
\end{table}

In Table \ref{tabTest}, we observe that PnO$_{\text{cp}}$ achieves the best overall rank (minimum value for the second column) and the highest average profit. The proposed method achieves superior performance when compared to the second-ranked approach (ProfLogit, 2.79 versus 4.42 in terms of average rank and 592$\textup{\euro}$ versus 411$\textup{\euro}$ in terms of average profit). PnO$_{\text{cp}}$ also statistically outperforms all the classifiers where traditional measures are used for validation. Notably, the two top-performing classifiers are PnO$_{\text{cp}}$ and ProfLogit, highlighting the benefits of profit-driven classification during the training process.

\subsection{Sensitivity analysis and discussion}\label{sec:sensanalysis}

In this section, we explore the sensitivity of the retention incentive $d$. This parameter is pivotal for the appropriate design of the campaign and for addressing potential risks. The models selected to showcase the results are the top-performing models: PnO${\text{cp}}$ and ProfLogit, as well as both MSP${\text{SVM}}$ and SVM, given that they consistently achieve the best classification among all other base algorithms in terms of profit.

\begin{figure}[ht!]
    \centering
    \includegraphics[width=0.75\textwidth]{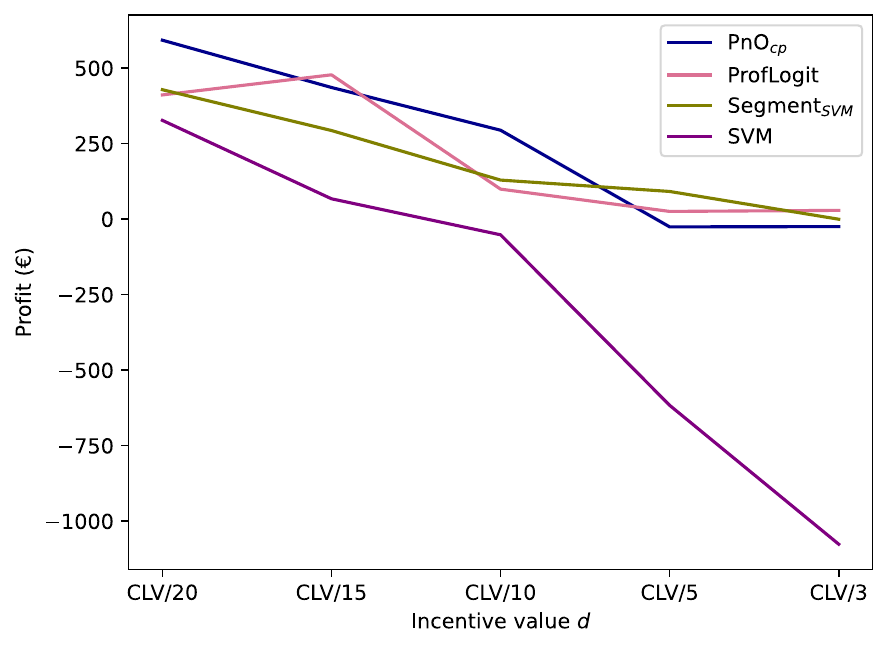}
    \caption{Profit over test set for an increasing value for the incentive $d$. The classification approaches that are shown are PnO$_{\text{cp}}$ (blue) ProfLogit (pink), MSP$_{\text{SVM}}$ (olive) and SVM (purple). The value shown is the mean taking into account the 12 different datasets.}
    \label{fig:profit_test}
\end{figure}

Figure \ref{fig:profit_test} illustrates the variation in terms profit for the mentioned methods as the value of $d$ increases, specifically for $d \in {\frac{\overline{CLV}}{20}, \frac{\overline{CLV}}{15}, \frac{\overline{CLV}}{10}, \frac{\overline{CLV}}{5}, \frac{\overline{CLV}}{3}}$. When the cost of the retention campaign is low, our model PnO$_{\text{cp}}$ exhibits clearly superior performance. However, as this cost increases, the profit-driven models tend to converge towards zero profit. This decline in performance might be attributed to the fact that, as retention campaign costs rise, fewer clients are valuable enough to cover the expenses. The intrinsic noise of the task makes it challenging to achieve a profitable campaign under these circumstances. Finally, we observe an important difference in performance between SVM, which is trained using traditional statistical measures, and the other models employing profit-driven metrics. This result clearly underscores the advantages of incorporating the downstream optimization task during the prediction process.

A similar analysis can be conducted by considering the profit achieved with the model relative to the maximum possible. As outlined in Section \ref{sec:summclas}, a normalized optimality gap can be computed for the profit earned over the test set using the formula:

\begin{equation}
 \frac{\sum_{i \in \mathcal{D}_\text{test}} c(z^*(y^i),y^i) - \sum_{i \in \mathcal{D}_\text{test}} c(z^*(\hat{y}^i),y^i)}{\sum_{i \in \mathcal{D}_\text{test}} c(z^*(y^i),y^i)}. 
\end{equation}

\begin{figure}[ht!]
    \centering
    \includegraphics[width=0.75\textwidth]{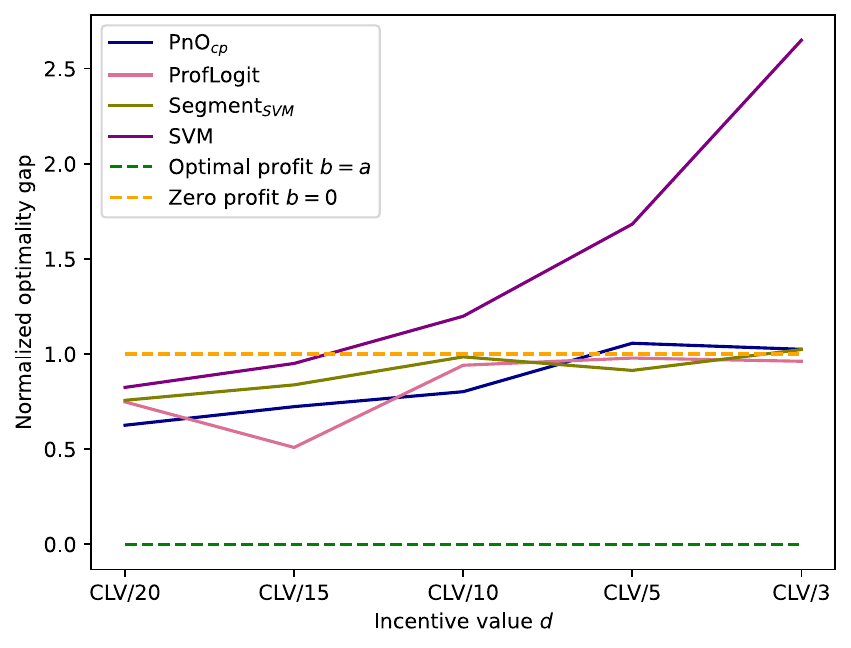}
    \caption{normalized optimality gap for an increasing value for the incentive $d$. The classification approaches that are shown are PnO$_{\text{cp}}$ (blue) ProfLogit (pink), MSP$_{\text{SVM}}$ (olive) and SVM (purple). Both obtaining the optimal profit (green) or zero profit (orange) are indicated with dashed lines. The value shown is the mean taking into account the 12 different datasets.}
    \label{fig:gap_test}
\end{figure}

This normalization is carried out to enhance interpretability. If a classifier achieves the maximum possible profit, then the gap is equal to 0. Conversely, if no profit is realized with the model, the gap equals 1. Values greater than one represent the worst-case scenario, indicating losses (i.e., the classifier produced a negative profit). Figure \ref{fig:gap_test} depicts this normalized optimality gap as the incentive $d$ increases for the selected methods. 

Figure \ref{fig:profit_test} reveals that PnO$_{\text{cp}}$ achieves the highest profit when $d=\frac{CLV}{20}$ and $d=\frac{CLV}{10}$, while ProfLogit is the model that gets closer to obtaining the optimal value when $d=\frac{CLV}{15}$. This result shows that including profit information during model training, as PnO$_{cp}$ and ProfLogit do, results in superior model performance.

\begin{figure}[ht!]
    \centering
    \includegraphics[width=\textwidth]{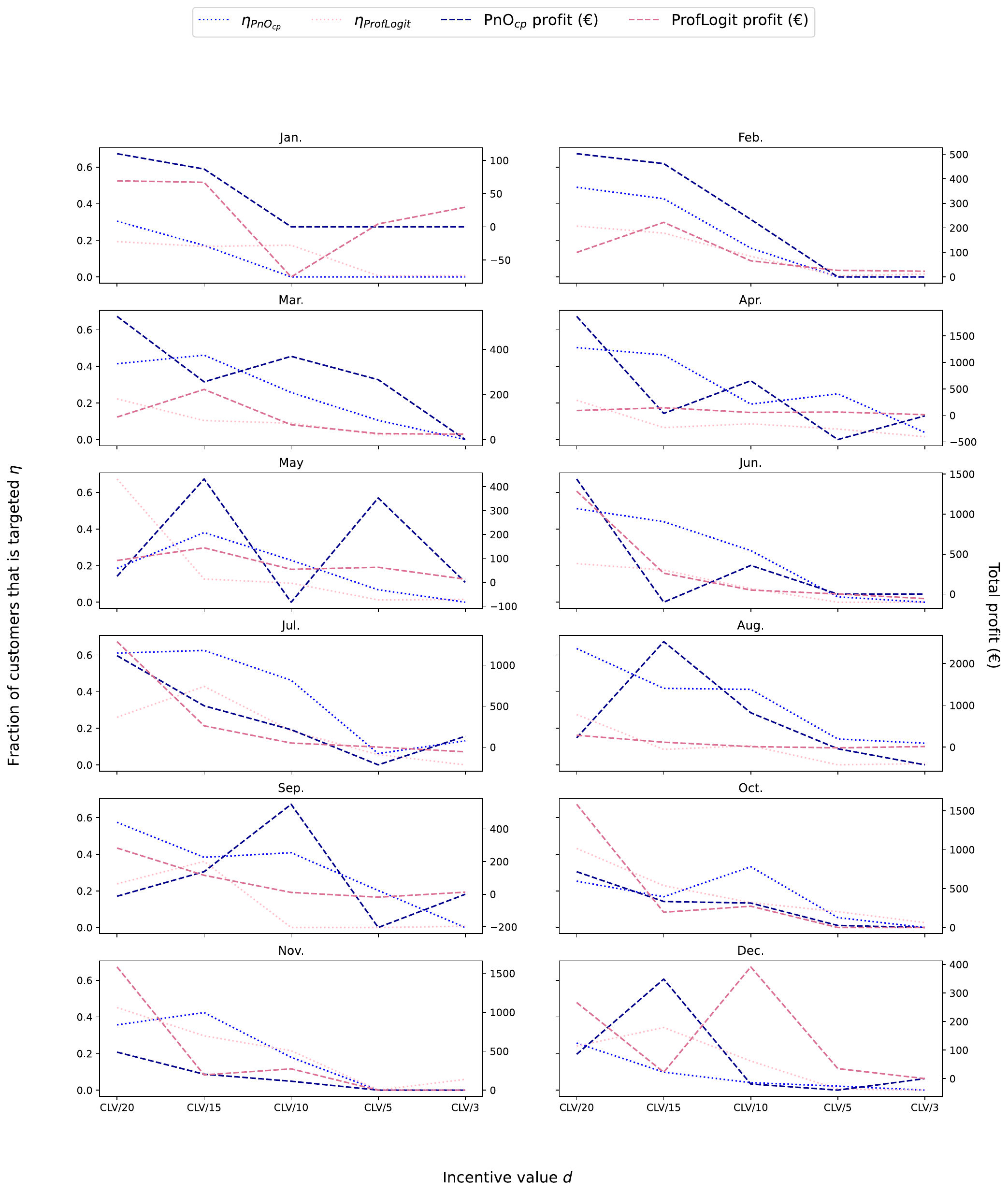}
    \caption{Fraction of customers that is targeted $\eta$ (left $y$ axis) and total profit (right $y$ axis) as incentive value $d$ increases ($x$ axis) for models PnO$_{cp}$ (blue) and ProfLogit (pink) in each of the 12 datasets.}
    \label{fig:frac_profit_test}
\end{figure}

Finally, Figure \ref{fig:frac_profit_test} presents the profit and the fraction of targeted customers $\eta$ for the various datasets and $d$ values. To simplify the interpretation of the results, we only present findings for the two top-performing models: PnO$_{\text{cp}}$ and ProfLogit. The goal of this analysis is to discern whether higher profits result from merely targeting a larger number of customers, or if the model accurately identifies a smaller subset of customers with higher CLVs.

For the first dataset in Figure \ref{fig:frac_profit_test} (January), for example, we observe that 
PnO$_{\text{cp}}$ achieves the highest profit while targeting the same fraction of customers as ProfLogit when $d=\frac{CLV}{15}$. Moreover, in December with $d=\frac{CLV}{15}$, the profit generated by PnO$_{\text{cp}}$ is significantly greater, even though it reaches out to less than half the number of customers that ProfLogit does. This findings underscore the advantages of accounting for individual CLVs rather than aggregating this information homogeneously. 

From our sensitivity analysis, we recommend examining multiple scenarios with varied incentive values. This approach simplifies the assessment of different campaign scenarios in relation to costs and the proportion of targeted clients, as illustrated in Figure \ref{fig:frac_profit_test}. Moreover, our comparison across models of different natures has revealed that relying solely on traditional statistical measures does not guarantee the optimization of the specific objective at hand --in our context, the total profit from a retention campaign--. This underscores the vital role of profit-driven methodologies and the importance of accounting for heterogeneous CLVs.

%% file: conclusions.tex
This study presents a novel profit-driven framework tailored for churn prediction. The main methodological contribution is the inclusion of heterogeneous CLVs during model training. Furthermore, our approach penalizes the indirect cost of excluding potential churners from campaigns, a misclassification error that conventional profit-based models disregard by attributing no cost. A predict-and-optimize approach is proposed for this challenge, considering the optimization goal (maximizing campaign profits) during the training step of the predictive model. As a result, customer inclusion in campaigns is determined not just by their churn probability but also by their individual lifetime values.

Experiments were performed on 12 different churn prediction datasets from a Chilean mutual fund company and our proposed method achieved the best overall rank and the highest average profit. This result shows the virtues of a modeling approach that account for individual-level CLVs typical of the investment sector and other industries. Moreover, the results highlight the superiority in performance of methods that maximize profit in the training step, as our model PnO$_{\text{cp}}$ and ProfLogit do. The enhanced performance of the former is likely attributed to its consideration of individual CLVs and its adept handling of prediction errors using the regret framework.

This study opens several opportunities for future developments. For example, an important challenge is how to properly address the stochasticity in the fraction $\gamma$ of the would-be churners that accept the incentive and remain with the company, an aspect not delved into in this study. Furthermore, our framework holds potential for customizing incentives $d_i$ for individual customers. Similar to personalized pricing, it is plausible that customers with higher CLVs might require more substantial incentives and superior offers. Unfortunately, the data in this study lacks specifics on retention campaigns, preventing an assessment of such initiatives for now. 

Another limitation of this study is the use of single-source datasets from the financial domain. This decision stems from the scarcity of publicly-available datasets that encompass individual CLVs. Nonetheless, a comprehensible experimental analysis based on 12 datasets is presented. In the future, we aim to explore other sectors where retention campaigns are prevalent, including telecommunications and banking.

%% file: mainArXiv.bbl
\begin{thebibliography}{10}
\providecommand{\url}[1]{#1}
\csname url@samestyle\endcsname
\providecommand{\newblock}{\relax}
\providecommand{\bibinfo}[2]{#2}
\providecommand{\BIBentrySTDinterwordspacing}{\spaceskip=0pt\relax}
\providecommand{\BIBentryALTinterwordstretchfactor}{4}
\providecommand{\BIBentryALTinterwordspacing}{\spaceskip=\fontdimen2\font plus
\BIBentryALTinterwordstretchfactor\fontdimen3\font minus
  \fontdimen4\font\relax}
\providecommand{\BIBforeignlanguage}[2]{{%
\expandafter\ifx\csname l@#1\endcsname\relax
\typeout{** WARNING: IEEEtran.bst: No hyphenation pattern has been}%
\typeout{** loaded for the language `#1'. Using the pattern for}%
\typeout{** the default language instead.}%
\else
\language=\csname l@#1\endcsname
\fi
#2}}
\providecommand{\BIBdecl}{\relax}
\BIBdecl

\bibitem{maldonado2021profit}
S.~Maldonado, G.~Dom{\'\i}nguez, D.~Olaya, and W.~Verbeke, ``Profit-driven
  churn prediction for the mutual fund industry: a multisegment approach,''
  \emph{Omega}, vol. 100, p. 102380, 2021.

\bibitem{tungare2023net}
N.~R. Tungare and A.~Jain, ``Net promoter score (nps) framework for improving
  customer loyalty in supermarket,'' \emph{The Online Journal of Distance
  Education and e-Learning}, vol.~11, no.~2, 2023.

\bibitem{fujo2022customer}
S.~W. Fujo, S.~Subramanian, M.~A. Khder \emph{et~al.}, ``Customer churn
  prediction in telecommunication industry using deep learning,''
  \emph{Information Sciences Letters}, vol.~11, no.~1, p.~24, 2022.

\bibitem{Neslin2006}
S.~Neslin, S.~Gupta, W.~Kamakura, J.~Lu, and C.~Mason, ``Defection detection:
  Measuring and understanding the predictive accuracy of customer churn
  models,'' \emph{Journal of Marketing Research}, vol.~43, no.~2, pp. 204--211,
  2006.

\bibitem{verbeke2017profit}
W.~Verbeke, B.~Baesens, and C.~Bravo, \emph{Profit driven business analytics: A
  practitioner's guide to transforming big data into added value}.\hskip 1em
  plus 0.5em minus 0.4em\relax John Wiley \& Sons, 2017.

\bibitem{verbraken2012novel}
T.~Verbraken, W.~Verbeke, and B.~Baesens, ``A novel profit maximizing metric
  for measuring classification performance of customer churn prediction
  models,'' \emph{IEEE transactions on knowledge and data engineering},
  vol.~25, no.~5, pp. 961--973, 2012.

\bibitem{MALDONADO2021113493}
S.~Maldonado, J.~Miranda, D.~Olaya, J.~V{\'a}squez, and W.~Verbeke,
  ``Redefining profit metrics for boosting student retention in higher
  education,'' \emph{Decision Support Systems}, vol. 143, p. 113493, 2021.

\bibitem{DEVRIENDT2021497}
F.~Devriendt, J.~Berrevoets, and W.~Verbeke, ``Why you should stop predicting
  customer churn and start using uplift models,'' \emph{Information Sciences},
  vol. 548, pp. 497--515, 2021.

\bibitem{LOPEZ2019190}
J.~L\'opez and S.~Maldonado, ``Profit-based credit scoring based on robust
  optimization and feature selection,'' \emph{Information Sciences}, vol. 500,
  pp. 190--202, 2019.

\bibitem{MALDONADO2020273}
S.~Maldonado, J.~L{\'o}pez, and C.~Vairetti, ``Profit-based churn prediction
  based on minimax probability machines,'' \emph{European Journal of
  Operational Research}, vol. 284, no.~1, pp. 273--284, 2020.

\bibitem{stripling2018profit}
E.~Stripling, S.~vanden Broucke, K.~Antonio, B.~Baesens, and M.~Snoeck,
  ``Profit maximizing logistic model for customer churn prediction using
  genetic algorithms,'' \emph{Swarm and Evolutionary Computation}, vol.~40, pp.
  116--130, 2018.

\bibitem{Oskarsdottir2018}
M.~{\'O}skard{\'o}ttir, B.~Baesens, and J.~Vanthienen, ``Profit based model
  selection for customer retention using individual customer lifetime values,''
  \emph{Big Data}, vol.~6, no.~1, pp. 53--65, 2018.

\bibitem{vanderschueren2022predict}
T.~Vanderschueren, T.~Verdonck, B.~Baesens, and W.~Verbeke,
  ``Predict-then-optimize or predict-and-optimize? an empirical evaluation of
  cost-sensitive learning strategies,'' \emph{Information Sciences}, vol. 594,
  pp. 400--415, 2022.

\bibitem{elmachtoub2022smart}
A.~N. Elmachtoub and P.~Grigas, ``Smart ``predict, then optimize'',''
  \emph{Management Science}, vol.~68, no.~1, pp. 9--26, 2022.

\bibitem{ahmad2019customer}
A.~K. Ahmad, A.~Jafar, and K.~Aljoumaa, ``Customer churn prediction in telecom
  using machine learning in big data platform,'' \emph{Journal of Big Data},
  vol.~6, no.~1, pp. 1--24, 2019.

\bibitem{Verbeke2012}
W.~Verbeke, K.~Dejaeger, D.~Martens, J.~Hur, and B.~Baesens, ``New insights
  into churn prediction in the telecommunication sector: A profit driven data
  mining approach,'' \emph{European Journal of Operational Research}, vol. 218,
  no.~1, pp. 211--229, 2012.

\bibitem{mandi2023decisionfocused}
J.~Mandi, J.~Kotary, S.~Berden, M.~Mulamba, V.~Bucarey, T.~Guns, and
  F.~Fioretto, ``Decision-focused learning: Foundations, state of the art,
  benchmark and future opportunities,'' 2023.

\bibitem{agrawal2019differentiable}
A.~Agrawal, B.~Amos, S.~Barratt, S.~Boyd, S.~Diamond, and J.~Z. Kolter,
  ``Differentiable convex optimization layers,'' \emph{Advances in neural
  information processing systems}, vol.~32, 2019.

\bibitem{amos2017optnet}
B.~Amos and J.~Z. Kolter, ``Optnet: Differentiable optimization as a layer in
  neural networks,'' in \emph{International Conference on Machine
  Learning}.\hskip 1em plus 0.5em minus 0.4em\relax PMLR, 2017, pp. 136--145.

\bibitem{blondel2020fast}
M.~Blondel, O.~Teboul, Q.~Berthet, and J.~Djolonga, ``Fast differentiable
  sorting and ranking,'' in \emph{International Conference on Machine
  Learning}.\hskip 1em plus 0.5em minus 0.4em\relax PMLR, 2020, pp. 950--959.

\bibitem{mandi2020interior}
J.~Mandi and T.~Guns, ``Interior point solving for lp-based prediction+
  optimisation,'' \emph{Advances in Neural Information Processing Systems},
  vol.~33, pp. 7272--7282, 2020.

\bibitem{wilder2019melding}
B.~Wilder, B.~Dilkina, and M.~Tambe, ``Melding the data-decisions pipeline:
  Decision-focused learning for combinatorial optimization,'' in
  \emph{Proceedings of the AAAI Conference on Artificial Intelligence},
  vol.~33, 2019, pp. 1658--1665.

\bibitem{poganvcic2019differentiation}
M.~V. Pogan{\v{c}}i{\'c}, A.~Paulus, V.~Musil, G.~Martius, and M.~Rolinek,
  ``Differentiation of blackbox combinatorial solvers,'' in \emph{International
  Conference on Learning Representations}, 2019.

\bibitem{sahoo2022backpropagation}
S.~S. Sahoo, A.~Paulus, M.~Vlastelica, V.~Musil, V.~Kuleshov, and G.~Martius,
  ``Backpropagation through combinatorial algorithms: Identity with projection
  works,'' in \emph{The Eleventh International Conference on Learning
  Representations}, 2022.

\bibitem{mandi2022decision}
J.~Mandi, V.~Bucarey, M.~M.~K. Tchomba, and T.~Guns, ``Decision-focused
  learning: through the lens of learning to rank,'' in \emph{International
  Conference on Machine Learning}.\hskip 1em plus 0.5em minus 0.4em\relax PMLR,
  2022, pp. 14\,935--14\,947.

\bibitem{mulamba2021contrastive}
M.~Mulamba, J.~Mandi, M.~Diligenti, M.~Lombardi, V.~B. Lopez, and T.~Guns,
  ``Contrastive losses and solution caching for predict-and-optimize,'' in
  \emph{30th International Joint Conference on Artificial Intelligence
  (IJCAI-21): IJCAI-21}.\hskip 1em plus 0.5em minus 0.4em\relax International
  Joint Conferences on Artificial Intelligence, 2021, pp. 2833--2840.

\bibitem{elmachtoub2020decision}
A.~N. Elmachtoub, J.~C.~N. Liang, and R.~McNellis, ``Decision trees for
  decision-making under the predict-then-optimize framework,'' in
  \emph{International Conference on Machine Learning}.\hskip 1em plus 0.5em
  minus 0.4em\relax PMLR, 2020, pp. 2858--2867.

\bibitem{Ketkar2021}
N.~Ketkar and J.~Moolayil, \emph{Automatic Differentiation in Deep
  Learning}.\hskip 1em plus 0.5em minus 0.4em\relax Berkeley, CA: Apress, 2021,
  pp. 133--145.

\bibitem{DUBEY202292}
S.~R. Dubey, S.~K. Singh, and B.~B. Chaudhuri, ``Activation functions in deep
  learning: A comprehensive survey and benchmark,'' \emph{Neurocomputing}, vol.
  503, pp. 92--108, 2022.

\bibitem{7410480}
K.~He, X.~Zhang, S.~Ren, and J.~Sun, ``Delving deep into rectifiers: Surpassing
  human-level performance on imagenet classification,'' in \emph{2015 IEEE
  International Conference on Computer Vision (ICCV)}, 2015, pp. 1026--1034.

\bibitem{mena2023exploiting}
G.~Mena, K.~Coussement, K.~W. De~Bock, A.~De~Caigny, and S.~Lessmann,
  ``Exploiting time-varying rfm measures for customer churn prediction with
  deep neural networks,'' \emph{Annals of Operations Research}, pp. 1--23,
  2023.

\bibitem{sun2019survey}
S.~Sun, Z.~Cao, H.~Zhu, and J.~Zhao, ``A survey of optimization methods from a
  machine learning perspective,'' \emph{IEEE transactions on cybernetics},
  vol.~50, no.~8, pp. 3668--3681, 2019.

\bibitem{bahnsen2014}
A.~C. Bahnsen, D.~Aouada, and B.~Ottersten, ``Example-dependent cost-sensitive
  logistic regression for credit scoring,'' in \emph{2014 13th International
  Conference on Machine Learning and Applications}, 2014, pp. 263--269.

\bibitem{hoppner2022instance}
S.~H\"oppner, B.~Baesens, W.~Verbeke, and T.~Verdonck, ``Instance-dependent
  cost-sensitive learning for detecting transfer fraud,'' \emph{European
  Journal of Operational Research}, vol. 297, no.~1, pp. 291--300, 2022.

\bibitem{chawla2002smote}
N.~V. Chawla, K.~W. Bowyer, L.~O. Hall, and W.~P. Kegelmeyer, ``Smote:
  synthetic minority over-sampling technique,'' \emph{Journal of artificial
  intelligence research}, vol.~16, pp. 321--357, 2002.

\bibitem{Demsar2006}
J.~Dem\v{s}ar, ``Statistical comparisons of classifiers over multiple data
  set,'' \emph{Journal of Machine Learning Research}, vol.~7, pp. 1--30, 2006.

\end{thebibliography}
